\documentclass[runningheads,a4paper]{llncs}

\usepackage{amssymb}
\usepackage{amsmath}
\usepackage{graphicx}
\usepackage{subfigure}
\usepackage{url}
\usepackage{color}
\usepackage[sort,comma]{natbib}
\usepackage[ruled,vlined]{algorithm2e} % for algorithms
\usepackage{multirow}

\urldef{\mailsa}\path|{muhammad.ghifary,bastiaan.kleijn,mengjie.zhang}@ecs.vuw.ac.nz|
\newcommand{\keywords}[1]{\par\addvspace\baselineskip
\noindent\keywordname\enspace\ignorespaces#1}

\begin{document}

\mainmatter  % start of an individual contribution

% first the title is needed
\title{Domain Adaptive Neural Networks for \\Object Recognition}

% a short form should be given in case it is too long for the running head
\titlerunning{Domain Adaptive Neural Networks for Object Recognition}

% the name(s) of the author(s) follow(s) next
%
% NB: Chinese authors should write their first names(s) in front of
% their surnames. This ensures that the names appear correctly in
% the running heads and the author index.
%
\author{Muhammad Ghifary%
%\thanks{Please note that the LNCS Editorial assumes that all authors have used
%the western naming convention, with given names preceding surnames. This determines
%the structure of the names in the running heads and the author index.}%
\and W. Bastiaan Kleijn\and Mengjie Zhang}

\authorrunning{Muhammad Ghifary, W. Bastiaan Kleijn, Mengjie Zhang}
% (feature abused for this document to repeat the title also on left hand pages)

% the affiliations are given next; don't give your e-mail address
% unless you accept that it will be published
\institute{School of Engineering and Computer Science\\
Victoria University of Wellington\\
PO Box 600, Wellington, New Zealand
\mailsa\\
%\mailsb\\
%\mailsc\\
%\url{http://www.springer.com/lncs}
}

%
% NB: a more complex sample for affiliations and the mapping to the
% corresponding authors can be found in the file "llncs.dem"
% (search for the string "\mainmatter" where a contribution starts).
% "llncs.dem" accompanies the document class "llncs.cls".
%

\toctitle{Domain Adaptive Neural Network for Object Recognition}
\tocauthor{Authors' Instructions}
\maketitle

\vspace{-1.0em}
\begin{abstract}
We propose a simple neural network model to deal with the domain adaptation problem in object recognition.
Our model incorporates the Maximum Mean Discrepancy (MMD) measure as a regularization in the supervised learning to reduce the distribution mismatch between the source and target domains in the latent space.
From experiments, we demonstrate that the MMD regularization is an effective tool to provide good domain adaptation models on both SURF features and raw image pixels of a particular image data set.
We also show that our proposed model, preceded by the denoising auto-encoder pretraining, achieves better performance than recent benchmark models on the same data sets.
This work represents the first study of MMD measure in the context of neural networks.
\\
\end{abstract}

\section{Introduction}
In learning-based computer vision, the probability distribution mismatch between the training and test samples is an essential problem to overcome for the success in real world scenarios.
For example, suppose we have an object recognizer learned from a training set containing objects with specific viewpoints, backgrounds, and transformations.
It is then applied to an environment with a similar object category, but different viewpoints, backgrounds, and transformations condition.
This situation might happen due to a lack of labeled data representing the target environment or insufficient knowledge regarding to the target condition.
A good recognition model on this setting can not be guaranteed if it is trained by using traditional learning techniques.

Methods to address the distribution mismatch have been investigated under the names of \emph{domain adaptation}\footnote{In this area, the term ``domain`` and ``probability distribution`` are considered to be identical.} and \emph{transfer learning}.
More specifically, given a training set $\{ \mathbf{x}^{(i)}_s, y^{(i)}_s \}_{i=1,...,n_s}$ and test set $\{ \mathbf{x}^{(j)}_t, y^{(j)}_t\}_{j=1,...,n_t}$ sampled from a distribution $\mathcal{D}_s$ and $\mathcal{D}_t$ respectively, 
the goal is to predict the target labels $y^{(j)}_t$ when $\mathcal{D}_s\neq \mathcal{D}_t$ and the information about $y^{(j)}_t$ is not sufficient. 
In recent years, many solutions to this problem have been proposed for computer vision applications~\citep{Saenko:2010aa,Gopalan:2011aa,Gong:2012aa,Long:2013aa} and natural language processing~\citep{Pan:2010aa,Daume-III:2009aa}.

In image recognition, the Office data set~\citep{Saenko:2010aa} has become a standard image set to evaluate the performance of domain adaptation models.
The standard evaluation protocol on this data set is based on using the SURF feature descriptor~\citep{Bay:2008aa} as inputs to the model.
However, the utilization of such a descriptor usually needs a careful engineering to get good discriminative features.
Furthermore, it may bring more complexity in the context of real time feature extraction processes.
It is therefore worthwhile to build good models without using any handcrafted feature descriptors.

Representation or feature learning provides a framework to reduce the dependency on manual feature engineering~\citep{Bengio:2012ab}.
Examples that can be considered as representation learning are Principal Component Analysis (PCA), Independent Component Analysis (ICA), Sparse Coding, Neural Networks, and Deep Learning.
In deep learning, the greedy layer-wise unsupervised training, which is known as the \emph{pretraining}, has played an important role for the success of deep neural networks~\citep{Bengio:2007aa,Erhan:2010aa}.
Although representation learning-based techniques have brought some successes over many applications, methods to address the distribution mismatch have not yet been well studied.

In this work, we propose a simple neural network model with good domain adaptation performance on raw image pixels.
More particularly, we utilize a non-parametric probability distribution distance measure, i.e, the Maximum Mean Discrepancy (MMD), as a regularization embedded in the supervised back-propagation training.
MMD is used to reduce the distribution mismatch between two hidden layer representations induced by samples drawn from different domains.
Despite its effectiveness, to our best knowledge, the use of MMD in the context of neural networks has not been investigated yet.
This work is therefore the first study to use MMD in neural networks. 
Specifically, we will investigate whether the MMD regularization can indeed improve the discriminative domain adaptation performance of neural networks.

\section{Preliminaries}
In this section, we will describe several tools related to our proposed method such as MMD measure, feed forward neural network, and denoising auto-encoder.
Some reviews about such tools in recent literature will be also included.

\vspace{-1.0em}
\subsection{Maximum Mean Discrepancy}
The Maximum Mean Discrepancy (MMD) is a measure of the difference between two probability distributions from their samples.
It is an effective criterion that compares distributions without initially estimating their density functions.
Given two probability distributions $p$ and $q$ on $\mathcal{X}$, MMD is defined as
\begin{eqnarray}
\label{eq:mmd}
	\mathcal{MMD}(\mathfrak{F}, p, q) = \mathrm{sup}_{f \in \mathfrak{F}} (\mathbb{E}_{x \sim p}\left[f(x) \right] - \mathbb{E}_{y \sim q}\left[f(y) \right]),
\end{eqnarray}
where $\mathfrak{F}$ is a class of functions $f : \mathcal{X} \rightarrow \mathbb{R}$. 
By defining $\mathfrak{F}$ as the set of functions of the unit ball in a universal Reproducing Kernel Hilbert Space (RKHS), denoted by $\mathcal{H}$, it was shown that $\mathcal{MMD}(\mathfrak{F}, p, q) = 0$ will detect any discrepancy between $p$ and $q$~\citep{Borgwardt:2006aa}.

Let $\{\mathbf{x}^{(i)}_s\}_{i=1,...,n_s}$ and $\{\mathbf{x}^{(j)}_t\}_{j=1,...,n_t}$ be data vectors drawn from distributions $\mathcal{D}_s$ and $\mathcal{D}_t$ on the data space $\mathcal{X}$, respectively.
Based on the fact that $f$ is in the unit ball in a universal RKHS, one may rewrite the empirical estimate of MMD as
\begin{eqnarray}
\label{eq:mmde}
\mathcal{MMD}_{\mathrm{e}} (\mathbf{x}_s, \mathbf{x}_t) = \left\| \frac{1}{n_s} \sum_{i=1}^{n_s} \phi(\mathbf{x}^{(i)}_s) - \frac{1}{n_t} \sum_{j=1}^{n_t} \phi(\mathbf{x}^{(j)}_t) \right\|_{\mathcal{H}},
\end{eqnarray}
where $\phi(\cdot) : \mathcal{X} \rightarrow \mathcal{H}$ is referred to as the \emph{feature space map}.

By casting (\ref{eq:mmde}) into a vector-matrix multiplication form, we come up with a \emph{kernelized} equation of the form~\citep{Borgwardt:2006aa}
\begin{eqnarray}
	\label{eq:mmde_kernel}
	\mathcal{MMD}_{\mathrm{e}}(\mathbf{x}_s, \mathbf{x}_t) &=&  
		\Bigg ( \frac{1}{n^2_s}\sum_{i=1}^{n_s} \sum_{j=1}^{n_s} k(\mathbf{x}^{(i)}_s, \mathbf{x}^{(j)}_s) 
		+ \frac{1}{n^2_t}\sum_{i=1}^{n_t} \sum_{j=1}^{n_t} k(\mathbf{x}^{(i)}_t, \mathbf{x}^{(j)}_t) \nonumber \\
		&-& \frac{2}{n_s n_t}\sum_{i=1}^{n_s} \sum_{j=1}^{n_s} k(\mathbf{x}^{(i)}_s, \mathbf{x}^{(j)}_t) \Bigg)^{\frac{1}{2}}
	 \\
	&=&  \left( \frac{\mathrm{Tr} \left(\mathbf{K}_{xss} \right)}{n^2_s} + 
		\frac{\mathrm{Tr} \left(\mathbf{K}_{xtt} \right)}{n^2_t} - 
		2 \frac{\mathrm{Tr} \left(\mathbf{K}_{xst} \right)}{n_s n_t} \right)^{\frac{1}{2}},
\end{eqnarray}
where 
$\left[\mathbf{K}_{x\bullet \bullet}\right]_{ij} = k(\mathbf{x}^{(i)}_{\bullet},\mathbf{x}^{(j)}_{\bullet})$
is the gram-matrix of all possible kernels in the data space.

In domain adaptation or transfer learning, MMD has been used to reduce the distribution mismatch between the source and target domain.
\citet{Pan:2009aa} proposed a PCA-based model referred to as Transfer Component Analysis (TCA) that used MMD to induce a subspace where the data distributions in different domains are closed to each other.
\citet{Long:2013aa} presented a Transfer Sparse Coding (TSC) that utilizes MMD in the encoding stage to match the distributions of the sparse codes.

Our work here adopts an idea of incorporating MMD into the learning algorithm similarly to TCA and TSC.
The difference is that we carry out the MMD regularization with respect to the supervised criterion while both TCA and TSC are unsupervised learning.
We expect that the MMD regularization embedded in the supervised training will induce better discriminative features.

\subsection{Feed Forward Neural Networks}
The Feed Forward Neural Network (FFNN) has been used extensively for solving many discrimative tasks during the past decades, including object recognition tasks.
The standard FFNN structure consists of three types of layer that are the input, hidden, and output layers with weighted \emph{inter-layer} connections.
The FFNN training corresponds to adjusting the connection weights with respect to a specific criterion.

Let us consider a single hidden layer neural network with $\mathbf{x} \in \mathbb{R}^{n_x}$, $\mathbf{h} \in \mathbb{R}^{n_h}$, and $\mathbf{o} \in \mathbf{R}^{n_o}$ as the visible, hidden, and output layers, respectively.
We denote $\mathbf{W}_1 \in \mathbb{R}^{n_x \times n_h}$ and $\mathbf{W}_2 \in \mathbb{R}^{n_h \times n_o}$ as the connection weights between the adjacent layers.
The FFNN can be written in the form of
\begin{eqnarray}
\label{eq:ffnn1}
	\mathbf{h} &=& \sigma_1(\mathbf{W}^{\top}_1 \mathbf{x} + \mathbf{b}), \\
	\mathbf{o} &=& \sigma_2(\mathbf{W}^{\top}_{2} \mathbf{h} + \mathbf{c}),
\end{eqnarray}
where $\mathbf{b} \in \mathbb{R}^{n_h}$ and $\mathbf{c} \in \mathbb{R}^{n_o}$ are the hidden and output units' biases, respectively. 
%In short, the FFNN can be simply described as
%\begin{eqnarray}
%\label{eq:ffnn2}
%	\mathbf{o} = g(\mathbf{x}) = \sigma_2(\mathbf{W}^{\top}_{2} \sigma_1(\mathbf{W}^{\top}_1 \mathbf{x} + \mathbf{b})+ \mathbf{c}).
%\end{eqnarray}

Note that both $\sigma_1 : \mathbb{R}^{n_h}\rightarrow \mathbb{R}^{n_h}$ and $\sigma_2 : \mathbb{R}^{n_o} \rightarrow \mathbb{R}^{n_o}$ are the non-linear activation functions.
In this work, we use the rectifier function approximated by the softplus function $\sigma_1(\mathbf{u})_j = \log(1+\exp(\mathbf{u}_j))$ and the softmax function $\sigma_2(\mathbf{v})_l = \frac{\exp(\mathbf{v}_l)}{\sum_{k} \exp(\mathbf{v}_k)}$, where $\mathbf{u} \in \mathbb{R}^{n_h}$ and $\mathbf{v} \in \mathbb{R}^{n_o}$.
The rectifier function $\sigma_1(\cdot)$ has been argued to be more biologically plausible than the logistic function~\citep{Glorot:2011ab}.
More importantly, several experimental works proved that the rectifier activation function can improve the performance of neural network models~\citep{Nair:2010aa}.
Furthermore, the use of the softmax function induces a probabilistic interpretation of the FFNN output.

Given the $n$ labeled training data $\{ \mathbf{x}^{i)}, \mathbf{y}^{(i)}\}_{i = 1,...,n}$, where $\mathbf{y} \in \{0,1\}^{n_o}$ represents the label with one active output node per class, 
the objective function of FFNN in the form of the empirical log-likelihood loss function is given as
\vspace{-0.5em}
\begin{eqnarray}
\label{eq:nn_loss}
J_{\mathrm{NN}} = -\frac{1}{n}\sum_{i=1}^{n} \sum_{k=1}^{l} \mathbf{y}^{(i)}_k \log \left( [g(\mathbf{x}^{(i)})]_k \right) 
\end{eqnarray}
which is typically minimized by the back-propagation algorithm.

\vspace{-0.5em}
\subsection{Denoising Auto-encoder}
An auto-encoder refers to an unsupervised neural network used for learning efficient codings.
In deep learning research, it is known as an effective technique for pretraining deep neural networks~\citep{Bengio:2007aa}.
In terms of the structure, the auto-encoder is very similar to the standard feed-forward neural network except that its output layer has an equal number of nodes as the input layer.
The objective of the auto-encoder is to reconstruct its own inputs by means of a reconstruction loss function.

A denoising auto-encoder (DAE) is a variant of the auto-encoder model that captures robust representations by reconstructing clean inputs given their noisy counterparts~\citep{Vincent:2010aa}.
Qualitatively, the use of several types of noise such as zero masking, Gaussian, and salt-and-pepper noises characterizes particular ``filters`` that correspond to the first hidden layer parameters~\citep{Vincent:2010aa}.
DAEs have been considered better than standard auto-encoders and comparable to restricted Boltzmann machines in the context of deep learning discriminative performance~\citep{Erhan:2010aa,Vincent:2010aa}.

In this work, we consider DAE as the pretraining stage of our proposed domain adaptive model.
Unlabeled images from both source and target domains are considered as inputs to the DAE pretraining.
We will investigate the effect with and without the DAE pretraining regarding to the domain adaptation performance.

\vspace{-0.5em}
\section{Domain Adaptive Neural Networks}
\vspace{-0.5em}
We propose a variant of the standard feed forward neural network that we refer to as the \emph{Domain Adaptive Neural Network} (DaNN).
This model incorporates MMD measure (\ref{eq:mmde}) as a regularization embedded in the supervised back-propagation training.
By using such a regularization, we aim to train the network parameters such that the supervised criterion is optimized and the hidden layer representations are encouraged to be invariant across different domains.

Given the labeled \emph{source} data $\{\mathbf{x}^{(i)}_s, \mathbf{y}^{(i)}_s\}_{i=1,...,n_s}$ and the unlabeled \emph{target} data $\{ \mathbf{x}^{(j)}_t \}_{j = 1,..., n_t}$, the loss function of a single layer DaNN is given by
\vspace{-0.2em}
\begin{eqnarray}
\label{eq:dann_loss}
	J_{\mathrm{DaNN}} = J_{\mathrm{NNs}}+ \mathbf{\gamma} \mathcal{MMD}^2_e(\mathbf{q}_s, \mathbf{\bar{q}}_t),
\end{eqnarray}
where $J_{\mathrm{NNs}} = -\frac{1}{n_s}\sum_{i=1}^{n_s} \sum_{k=1}^{l} ([\mathbf{y}^{(i)}_{s}]_k \log([f(\mathbf{x}_s^{(i)})]_k) )$ is the same loss function as shown in (\ref{eq:nn_loss}) but applied only over the source data,
$\mathbf{q}_s = \mathbf{W}^{\top}_1 \mathbf{x}_s + \mathbf{b}$, 
$\mathbf{\bar{q}}_t = \mathbf{W}^{\top}_1 \mathbf{x}_t + \mathbf{b}$ are the linear combination outputs before the activation, and 
$\gamma$ is the regularization constant controlling the importance of MMD contribution to the loss function.

To minimize ($\ref{eq:dann_loss}$), we need the gradient of $J_{\mathrm{DaNN}}$.
While computing the gradient of $J_{\mathrm{NNs}}$ over $\{ \mathbf{W}_1$,$\mathbf{b}, \mathbf{c} \}$ is trivial,
computing the gradient of $\mathcal{MMD}^2_e(\mathbf{q}_s, \mathbf{\bar{q}}_t)$ depends on the choice of the kernel function. 
We choose the Gaussian kernel, which is considered as a universal kernel~\citep{Steinwart:2002aa}, as the kernel function of the form
$k_G(\mathbf{x}, \mathbf{y}) = \exp \left(- \frac{\| \mathbf{x} - \mathbf{y} \|^{2}}{2 s^2} \right)$,
where $s$ is the standard deviation.

We can rewrite the $\mathcal{MMD}^{2}_e(\cdot, \cdot)$ function $(\ref{eq:dann_loss})$ in terms of the Gaussian kernel by a matrix-vector form.
Let us denote the sample vectors {\small $\mathbf{\tilde{x}^{(i)}}_s = \begin{bmatrix} 1 \\ \mathbf{x}^{(i)}_s\end{bmatrix} \in \mathbb{R}^{(d + 1)}, \forall i=1,..., n_s$} and {\small $\mathbf{\tilde{x}^{(j)}}_t = \begin{bmatrix} 1 \\ \mathbf{x}^{(j)}_t\end{bmatrix} \in \mathbb{R}^{(d + 1)}, \forall j=1,..., n_t$}.
The additional element of 1 in each sample is utilized to incorporate the computation with the biases.
Let us define the parameter matrices {\small $\mathbf{U}_1 = \begin{bmatrix} \mathbf{b}^{\top} \\ \mathbf{W}_1\end{bmatrix} \in \mathbb{R}^{(d + 1) \times k}$} and {\small $\mathbf{U}_2 = \begin{bmatrix} \mathbf{c}^{\top} \\ \mathbf{W}_2\end{bmatrix} \in \mathbb{R}^{(k + 1) \times l}$}.
Hence, the $\mathcal{MMD}^{2}_e(\cdot, \cdot)$ function can be rewritten as
\vspace{-1.0em}
{\small \begin{eqnarray}
	\label{eq:mmd2e}
	\mathcal{MMD}^{2}_e (\mathbf{U}^{\top}_1 \mathbf{X}_s, \mathbf{U}^{\top}_1 \mathbf{X}_t) 
&=& 
\frac{1}{n^2_s} \sum_{i,j=1}^{n_s} \exp \left( - \frac{(\mathbf{x}^{(i)}_s - \mathbf{x}^{(j)}_s)^{\top} \mathbf{U}_1  \mathbf{U}^{\top}_1  ( \mathbf{x}^{(i)}_s - \mathbf{x}^{(i)}_s ) }{2s^2}\right) \nonumber \\
&+&  \frac{1}{n^2_t} \sum_{i,j=1}^{n_t} \exp \left( - \frac{(\mathbf{x}^{(i)}_t - \mathbf{x}^{(j)}_t)^{\top} \mathbf{U}_1  \mathbf{U}^{\top}_1  ( \mathbf{x}^{(i)}_t - \mathbf{x}^{(i)}_t ) }{2s^2}\right) \nonumber \\
&-&  \frac{2}{n_s n_t} \sum_{i,j=1}^{n_s, n_t} \exp \left( - \frac{(\mathbf{x}^{(i)}_s - \mathbf{x}^{(j)}_t)^{\top} \mathbf{U}_1  \mathbf{U}^{\top}_1  ( \mathbf{x}^{(i)}_s - \mathbf{x}^{(i)}_t ) }{2s^2}\right).
\end{eqnarray} }
\vspace{-1em}

Let {\small $G_{\bullet \bullet}(i,j)$} be the gradient of {\small $k_G(\mathbf{U}^{\top}_1 \mathbf{x}^{(i)}_{\bullet}, \mathbf{U}^{\top}_1 \mathbf{x}^{(j)}_{\bullet})$}, where the symbol $\bullet$ can be either $s$ or $t$, with respect to $\mathbf{U}_1$.
Then, $G_{\bullet \bullet}(i,j)$ takes the form
\vspace{-0.8em}
{\small \begin{eqnarray}
\label{eq:grad_gaussiankernel}
G_{\bullet \bullet}(i,j) = - \frac{1}{s^2} k_G(\mathbf{x}^{(i)}_{\bullet}, \mathbf{x}^{(j)}_{\bullet}) 
(\mathbf{x}^{(i)}_{\bullet} - \mathbf{x}^{(j)}_{\bullet} ) (\mathbf{x}^{(i)}_{\bullet} - \mathbf{x}^{(j)}_{\bullet} )^{\top} \mathbf{U}_1.
\end{eqnarray}
}
Now it is straightforward to see that the gradient of {\small $\mathcal{MMD}^{2}_e (\mathbf{U}^{\top}_1 \mathbf{X}_s, \mathbf{U}^{\top}_1 \mathbf{X}_t)$} w.r.t $\mathbf{U}_1$ ($\frac{\partial \mathcal{M}^2_{st}}{\partial \mathbf{U}_1 }$ for short) is given by
\vspace{-1.0em}
{\small \begin{eqnarray}
\label{eq:grad_mmd}
\frac{\partial \mathcal{M}^2_{st}}{\partial \mathbf{U}_1} = \frac{1}{n^2_s} \sum_{i,j=1}^{n_s} G_{ss}(i,j)+ \frac{1}{n^2_t} \sum_{i,j=1}^{n_t} G_{tt}(i,j)- \sum_{i,j=1}^{n_s, n_t} \frac{2}{n_s n_t} G_{st}(i,j).
\end{eqnarray}}
\vspace{-1em}

The main reason for choosing the Gaussian kernel is that it has been well studied and proven to make MMD useful in practice~\citep{Gretton:2012aa}.
Furthermore, it is worth noting that MMD here is applied to linear combination outputs before we put on the non-linear activation function.
This means that MMD provides a biased estimate with respect to an actual distribution discrepancy of the hidden representations.
However, since we use the rectifier activation function that is close to linear, 
we expect that the measure in $(\ref{eq:mmd2e})$ would be able to produce good approximation of the true distribution discrepancy.

In the implementation, we separate the minimization of $J_{\mathrm{NNs}}$ and $\mathcal{MMD}^2_e(\cdot, \cdot)$ into two steps.
Firstly, $J_{\mathrm{NNs}}$ is minimized using a \emph{mini-batched} stochastic gradient descent with respect to  $\mathbf{U}_1$ update.
The mini-batched setting has become a standard practice in neural network training to establish a compromise between speed and accuracy.
Then, $\mathcal{MMD}^2_e(\cdot, \cdot)$ is minimized by re-updating $\mathbf{U}_1$ with respect to the gradient (\ref{eq:grad_mmd}).
The latter step is accomplished by a \emph{full-batched} gradient descent.
The detail of this procedure are summarized in Algorithm \ref{alg:dann}.

\scalebox{0.8}{
\begin{algorithm}[H]
 \SetAlgoLined % For v3.9
 %\SetAlgoLined % For previous releases [?]
%\tcc{ Train a deep hybrid network in a purely unsupervised, with the greedy layer-wise procedure in which the bottom layer is trained using auto-encoder learning with sparse regularization, and the upper layers are trained using stacked RBMs
%}
 \KwData{
\\ $\mathbf{U}_1 \in \mathbb{R}^{(d+1) \times k}$ and $\mathbf{U}_2 \in \mathbb{R}^{(k+1) \times l}$ are the weight-bias matrices in the first and second layers, respectively.
\\ $\mathbf{h} \in \mathbb{R}^k$ is the hidden layer vector.
\\ $\mathbf{o} \in \mathbb{R}^l$ is the output layer vector.
\\ $\alpha$, $\gamma$ are the learning rate and the MMD regularization constant.
 }
% \KwResult{how to write algorithm with \LaTeX2e }
 \Begin{
	\begin{enumerate}
	\item Initialize $\mathbf{U}_1$ and $\mathbf{U}_2$ with small random real values\;
	\item Update $\mathbf{U}_2$ and $\mathbf{U}_1$ using the batched stochastic gradient descent by the standard forward - backward pass w.r.t. $J_{\mathrm{NNs}}$\;
	\item Update $\mathbf{U}_1$ by the offline gradient descent as follows
		\begin{eqnarray}
			\mathbf{U}_1(t) := \mathbf{U}_1(t-1) - \alpha \gamma \frac{\partial \mathcal{M}^2_{st}}{\partial \mathbf{U}_1} \nonumber
		\end{eqnarray}
	\item Repeat Steps 2 and 3 until the end of the epoch\;
	\end{enumerate}
 }
 \caption{The DaNN supervised back-propagation algorithm.}
\label{alg:dann}
\end{algorithm}
}

\section{Experiments and Analysis}
\vspace{-0.5em}
We evaluated our proposed method in the context of object recognition over several domain mismatches.
We first compared the DaNN to baselines and other recent domain adaptation methods.
The results in terms of the recognition accuracy represented by the mean and standard deviation over 30 independent runs are then reported.
At last, we investigated the effect of the MMD regularization by measuring the difference of the first hidden layer activations between one domain to another domain.
\vspace{-1.0em}
\subsection{Setup}
\label{sec:ex_setup}
\begin{figure}
	\centering
	\includegraphics[width=3in,height=1.3in]{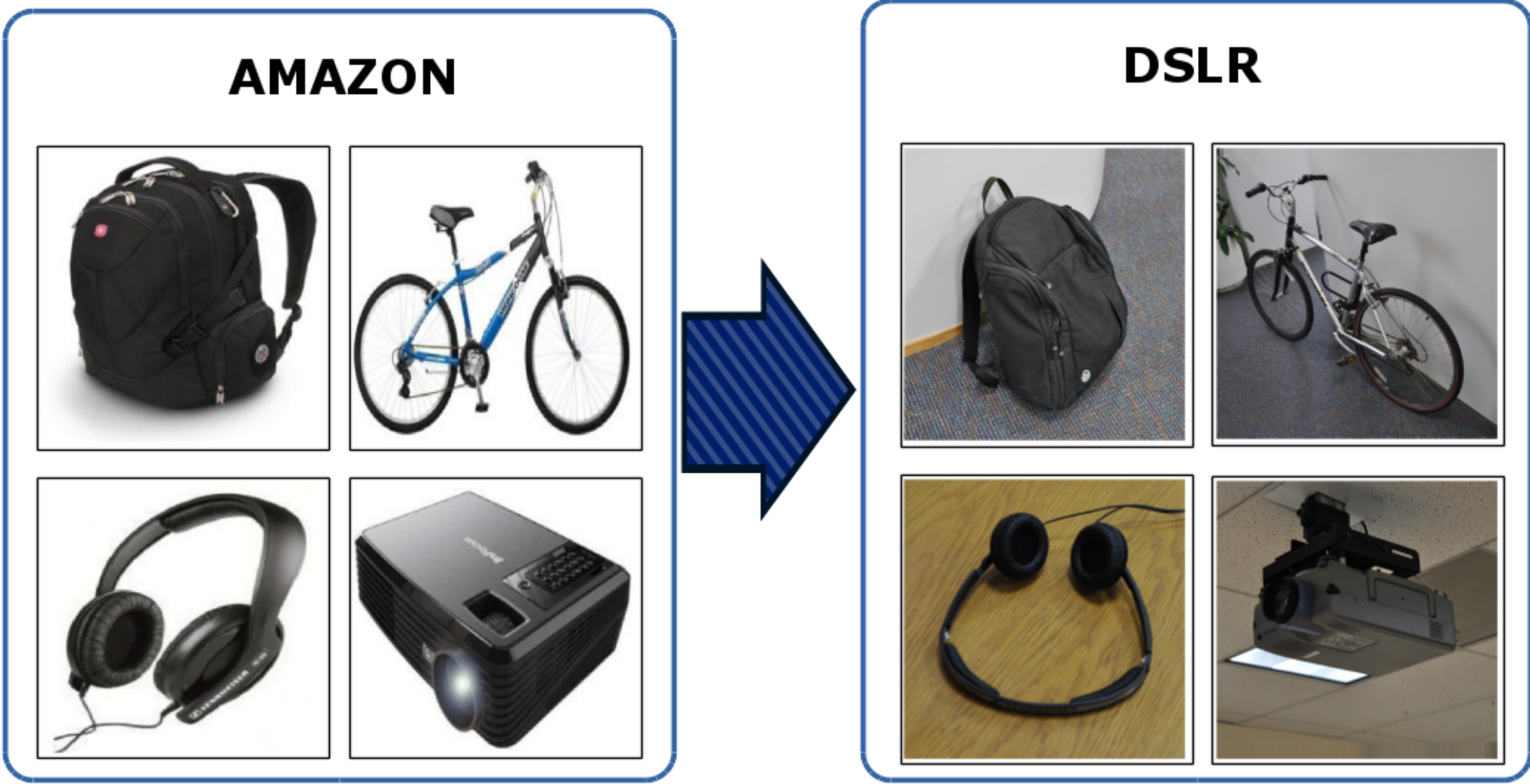}
	\vspace{-0.5em}
	\caption{The Office data set~\citep{Saenko:2010aa} samples from \emph{amazon} and \emph{dslr} domains.}
	\label{fig:office_samples}
\end{figure}

Our experiments used the Office data set~\citep{Saenko:2010aa} that contains images of 31 object  classes from three different domains: \emph{amazon}, \emph{webcam}, and \emph{dslr}.
In \emph{amazon}, the images contain a single centered object, while for the others the images were acquired in unconstrained settings with some variations such as lighting and background changes.
Here we only used 10 object classes following the protocol designed by \citet{Gong:2012aa}, which ends up with 1410 instances in total.
The number of images for \emph{amazon}, \emph{webcam}, and \emph{dslr}, respectively, are 958, 295, and 157.
\emph{Webcam} and \emph{dslr} are known to be more similar to each other based on the Rank of Domain (ROD) measure~\citep{Gong:2012aa}.
Examples of the Office images can be seen in Figure~\ref{fig:office_samples}.
\vspace{-1.0em}

The DaNN model used in the experiments has only one hidden layer, i.e., a shallow network of 256 hidden nodes.\footnote{This number was to obtain dimensionality reduction.
We tried other values such as 100, 300, and 500. Eventually, the number of 256 hidden nodes gave us the best performance among other values.
}
The input layer of the DaNN can be either raw pixels or SURF features.
The output layer contains ten nodes corresponding to the ten classes.

In all our experiments, we used the parameter setting for the supervised back-propagation learning specified in Table~\ref{tab:params}.
Note that we employed the \emph{dropout} regularization introduced by \citet{Hinton:2012aa}, the regularization of which randomly omits a hidden node for each training case with a certain probability.
It has been proven to produce better performance in the sense of reducing the overfitting if a neural network is trained from a small training set.
\vspace{-0.5em}
\begin{table}
	\centering
	\caption{The standard parameter setting of the DaNN.}	
	\vspace{-0.5em}
	\begin{tabular}{l || c}
	\hline
	Learning rate ($\alpha$) & 0.02 \\ 
	Iterations &900 \\
	Momentum & 0.05  \\
	L2 weight regularization & $0.003$ \\
	Dropout fraction & $0.5$ \\
	\hline
	\end{tabular}
	\label{tab:params}
\end{table}
\vspace{-1.0em}

For the MMD regularization, we set the standard deviation $s$ of the Gaussian kernel by the following calculation: $s = \sqrt{\frac{MSD}{2}}$~\citep{Baktashmotlagh:2013aa}, where $MSD$ is the median squared distance between all source samples. 
The MMD regularization constant $\gamma$ was set to be sufficiently large ($\gamma = 10^3$) to accommodate small values of (\ref{eq:grad_mmd}) compared to $\frac{J_{\mathrm{NNs}}}{\mathbf{U}_1}$ for each iteration.

We conducted six domain shift settings, each of which is a domain pair, based on three domains originated from the Office data set ($A \rightarrow W$, $W \rightarrow A$, $A \rightarrow D$, $D \rightarrow A$, $W \rightarrow D$, and $D \rightarrow W$).
The evaluation was divided into two settings: 1) \emph{unsupervised} adaptation, and 2) \emph{semi-supervised} adaptation.
The unsupervised adaptation corresponds to the setting when we can use both labeled images from the source domain and unlabeled images from the target domain during the training, but no labels from the target domain are incorporated.
In the semi-supervised adaptation, we incorporate a few labeled images from the target domain as additional training images. 
First three images per object category from the target domain are selected.
Differently from what was conducted in the initial work~\citep{Saenko:2010aa}, we used all labeled images from the source domain instead of randomly sampled from it.

The performance of our model was then compared to SVM-based baselines, two existing domain adaptation methods, and a simple neural network as follows:\\
\textbf{L-SVM}: an SVM~\citep{Cortes:1995aa} model with a linear kernel  that was applied to the original features.\footnote{\url{http://www.csie.ntu.edu.tw/~cjlin/liblinear}} \\
\textbf{L-SVM + PCA}: the same model as the L-SVM but preceded by PCA to reduce feature dimensionality. \\
\textbf{GFK}~\citep{Gong:2012aa}:
the Geodesic Flow Kernel approach by considering an infinite number of intermediate subspaces between the source and target domains followed by k-NN classification.\footnote{Here we used the subspaces constructed by PCA only} \\
\textbf{TSC}~\citep{Long:2013aa}:
the Transfer Sparse Coding technique based on the combination of the graph regularized sparse coding,  the MMD regularization, and the logistic regression.\footnote{\url{http://learn.tsinghua.edu.cn:8080/2011310560/long.html}}\\
\textbf{NN}:
a single layer neural network with the same structure and parameter setting (Table~\ref{tab:params}) used in our DaNN, but without the MMD regularization.\footnote{It is basically Algorithm \ref{alg:dann} without Step 3.}

\subsection{Results on SURF Features}
We first investigated the performance of our model on the standard image features provided by \citet{Gong:2012aa}.
Briefly, the image features were acquired by first utilizing the SURF descriptor on resized and grayscaled images to detect local scale-invariant interest points.
It was then followed by encoding the data points into 800-bin histograms using a codebook trained from a subset of \emph{amazon} images~\citep{Saenko:2010aa}. 
The final features were then normalized and z-scored to have zero mean and unit variance.
We conducted the unsupervised setting evaluation with the results shown in Table~\ref{tab:surf_uns}.

We found that DaNN and TSC have better performance than the other approaches on these standard features.
More specifically, DaNN performs well when there is the \emph{amazon} set in a particular domain pairs.
In the case of \emph{webcam}-\emph{dslr} shifts, the TSC, which has not been tested on the Office dataset in the previous work, is surprisingly the best model.
Despite its effectiveness, TSC has longer feature extraction time than, for example, neural network-based approaches so that it is less efficient in real world situation.
We also noted that the GFK, which incorporates multiple intermediate subspaces, fails to surpass the baselines in several cases. 
This indicates that the projection onto the subspaces generated by GFK is insufficient to reduce the domain mismatch.

%\begin{table}[h]
%\centering
%	\caption{The \emph{unsupervised} domain adaptation performance on the Office dataset (a : \emph{amazon}, w : \emph{webcam}, d : \emph{dslr}) using the SURF as inputs.}
%	\scalebox{0.9}{
%	\begin{tabular}{ c || c | c | c | c | c | c}
%		\hline
%		Methods & a $\rightarrow$ w & w $\rightarrow$ a& a $\rightarrow$ d & d $\rightarrow$ a  &  w $\rightarrow$ d & d $\rightarrow$ w \\ \hline
%		L-SVM & $24.07 \pm 0.00$ &  $35.80 \pm 0.00 $ & $28.03 \pm 0.00$  & $32.67 \pm 0.00$ & $77.71 \pm 0.00 $ &  $77.96 \pm 0.00$\\ \hline
%		PCA + L-SVM & $34.92 \pm 0.00 $ & $34.76 \pm 0.00$ &  $35.03 \pm 0.00$ & $32.15 \pm 0.00$ & $63.69 \pm 0.00 $  & $65.42 \pm 0.00$   \\ \hline
%		GFK~\cite{Gong:2012aa} &  $38.98 \pm 0.00$&$ 29.75 \pm 0.00$  &  $36.31 \pm 0.00$ & $31.84 \pm 0.00$ &  $80.25 \pm 0.00 $ & $75.59 \pm 0.00$  \\ \hline
%		TSC~\cite{Long:2013aa} & {\color{red}$47.42 \pm 1.69$}  & {\color{red}$39.12 \pm 0.38$} & $46.18 \pm 1.42$ & {\color{red} $41.55 \pm 0.80$}&  {\color{red}$93.63 \pm 0.52$} & {\color{red}$93.53 \pm 0.61$} \\ \hline
%		NN& $44.41 \pm ...$ & $37.27 \pm ...$  & $47.77 \pm ... $ & $34.76 \pm ...$ & $81.53 \pm ... $  & $78.89 \pm ...$ \\ \hline
%%		Pretrained NN + dropout &  &  & & &  &   \\ \hline
%		\textbf{DaNN} & $45.42 \pm ... $ &  $38.73 \pm ...$ &   {\color{red}$49.04 \pm ... $} & $38.10 \pm ...$& $83.44 \pm ...$ &  $81.02 \pm ...$\\ \hline
%%		\textbf{Pretrained DaNN + dropout} &  &   &   &  &  &  \\ \hline
%	\end{tabular}
%	}
%\end{table}
\vspace{-1.0em}
\begin{table}[!htb]
\centering
	\caption{The unsupervised setting performances on the Office data set ($A$ : \emph{amazon}, $W$ : \emph{webcam}, $D$ : \emph{dslr}) for each domain pair using SURF-based features as inputs. 
	Each column header starting from the second column indicates one domain pair, e.g., $A \rightarrow W$ represents the \emph{amazon} and \emph{webcam} as the training and test sets.
	}
	\scalebox{0.9}{
	\begin{tabular}{ c || c | c || c | c || c | c}
		\hline
		Methods & $A \rightarrow W$ & $W \rightarrow A$& $A \rightarrow D$ & $D \rightarrow A$  &  $W \rightarrow D$ & $D \rightarrow W$\\ \hline \hline
		L-SVM & $24.1 \pm 0.0$ &  $35.8 \pm 0.0 $ & $28.0 \pm 0.0$  & $32.7 \pm 0.0$ & $77.7 \pm 0.0 $ &  $78.0 \pm 0.0$\\ \hline
		PCA + L-SVM & $34.9 \pm 0.0 $ & $34.8 \pm 0.0$ &  $35.0 \pm 0.0$ & $32.2 \pm 0.0$ & $63.7 \pm 0.0$  & $65.4 \pm 0.0$   \\ \hline
		GFK~\cite{Gong:2012aa} &  $39.0 \pm 0.0$&$ 29.8 \pm 0.0$  &  $36.3 \pm 0.0$ & $31.8 \pm 0.0$ &  $80.3 \pm 0.0 $ & $75.6 \pm 0.0$  \\ \hline
		TSC~\cite{Long:2013aa} & \underline{$\mathbf{47.4 \pm 1.7}$}  & \underline{$\mathbf{39.1 \pm 0.4}$} & $46.2 \pm 1.4$ & \underline{$\mathbf{41.6 \pm 0.8}$}&  \underline{$\mathbf{93.6 \pm 0.5}$} & \underline{$\mathbf{93.5 \pm 0.6}$} \\ \hline
		NN& $44.4 \pm 0.6$ & $37.3 \pm 0.1$  & $47.8 \pm 0.9 $ & $34.8 \pm 0.2$ & $81.5 \pm 0.0 $  & $78.9 \pm 0.0$ \\ \hline
		\textbf{DaNN} & \underline{$\mathbf{45.4 \pm 0.8}$} &  \underline{$\mathbf{38.7 \pm 0.2}$} &  \underline{$\mathbf{49.0 \pm 0.7 }$} & $38.1 \pm 0.3$& $83.4 \pm 0.0$ &  $81.0 \pm 0.0$\\ \hline
	\end{tabular}
	\label{tab:surf_uns}
	}
\end{table}
\subsection{Results on Raw Pixels}
We also conducted the evaluation against the raw pixels of the Office images.
Previous works on the Office image set were mostly done using the SURF-based features. 
It is worth investigating the performance on the Office raw pixels directly since good models on raw pixels are preferable in the sense of reducing the needs of handcrafted feature extractors. 
We first converted the pixels of the Office images in 2D RGB values into grayscaled pixels and resized them into a dimension of $28 \times 28$. 
They were then z-scored to have zero mean and unit variance.

\subsection*{Domain Adaptation Setting}
\begin{figure}[htp]
        \centering
        \begin{center}
	    \subfigure[\emph{amazon}-\emph{webcam}]{\label{fig:a_w}\includegraphics[width=1.42in]{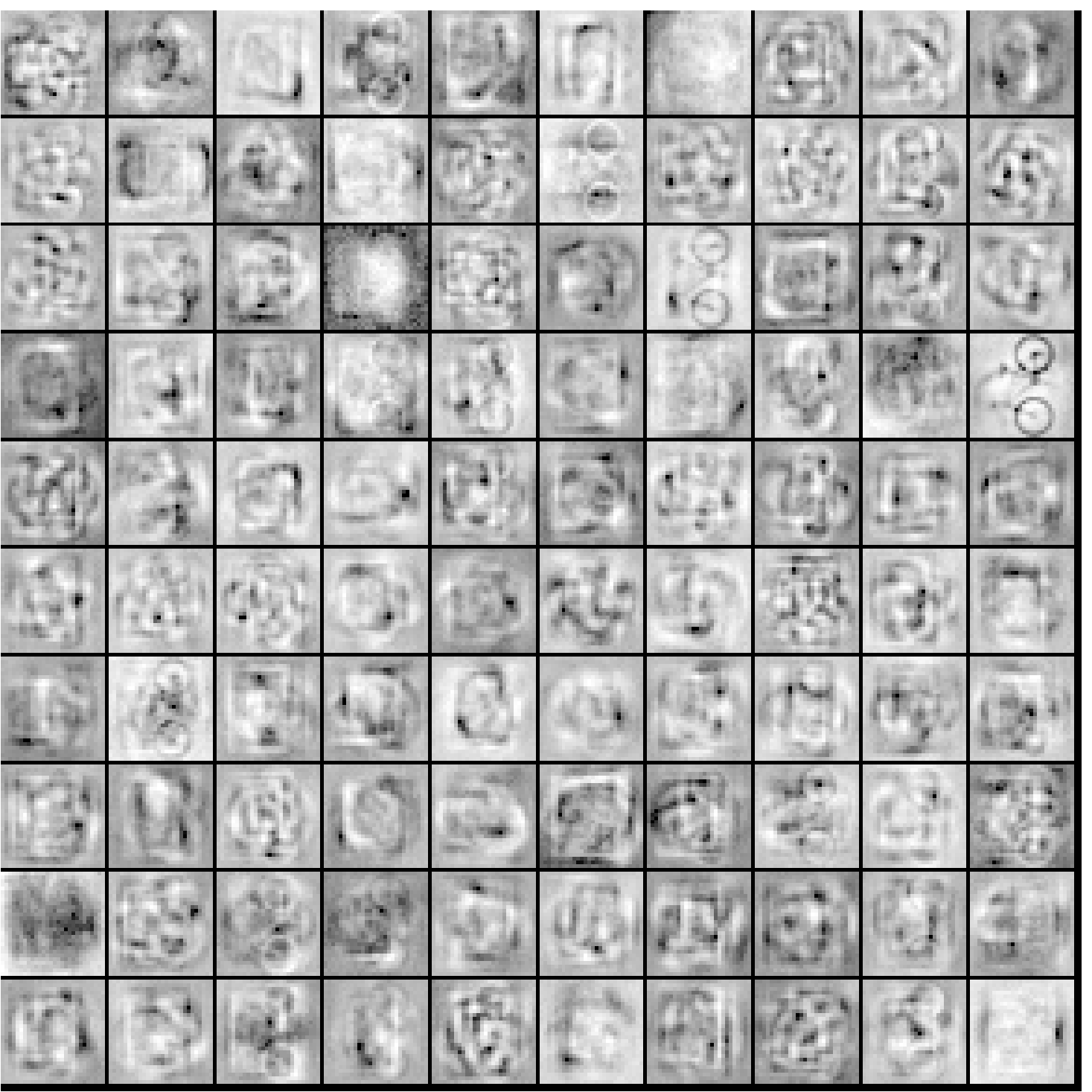}} \quad 
	    \subfigure[\emph{amazon}-\emph{dslr}]{\label{fig:a_d}\includegraphics[width=1.42in]{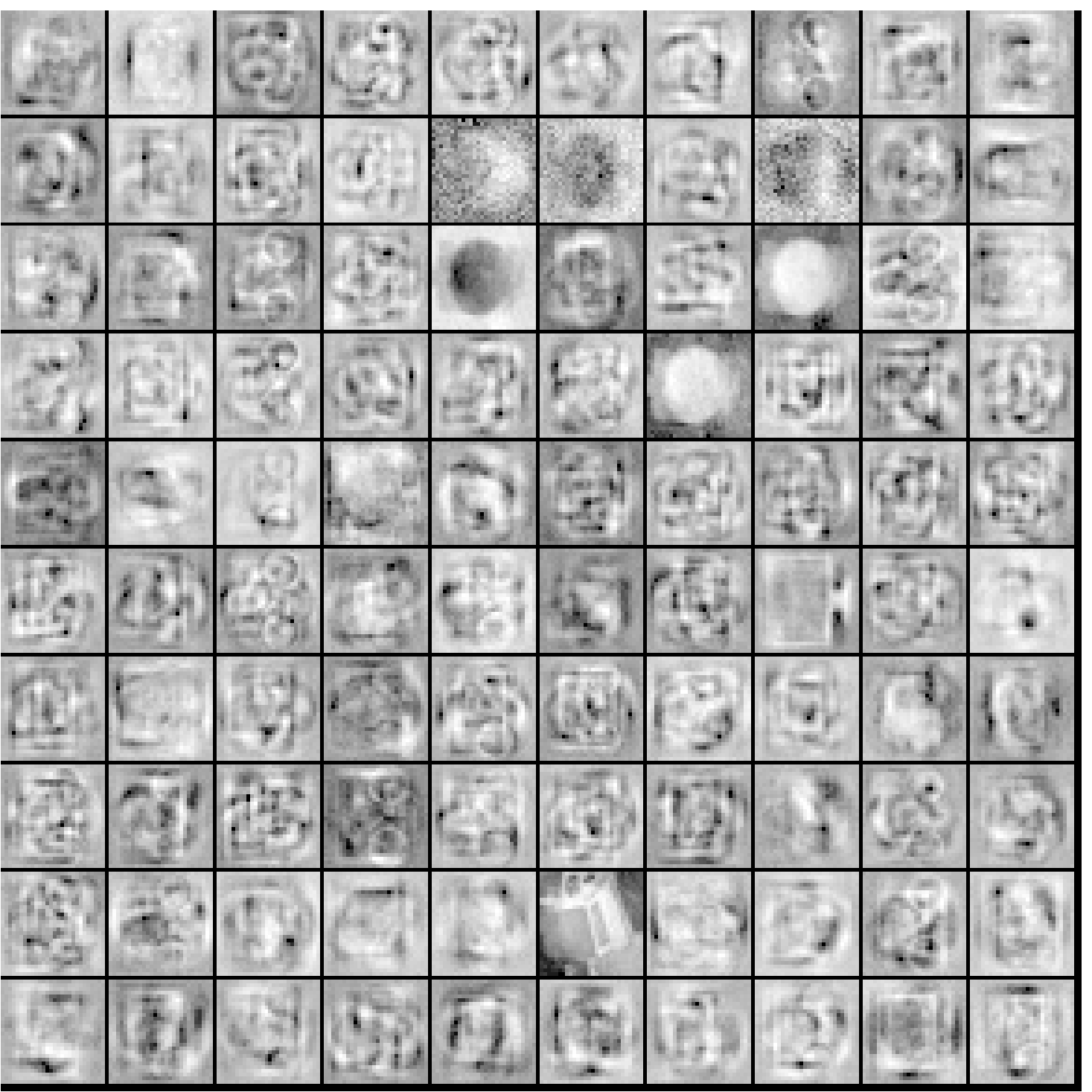}} \quad
	    \subfigure[\emph{webcam}-\emph{dslr}]{\label{fig:w_d}\includegraphics[width=1.42in]{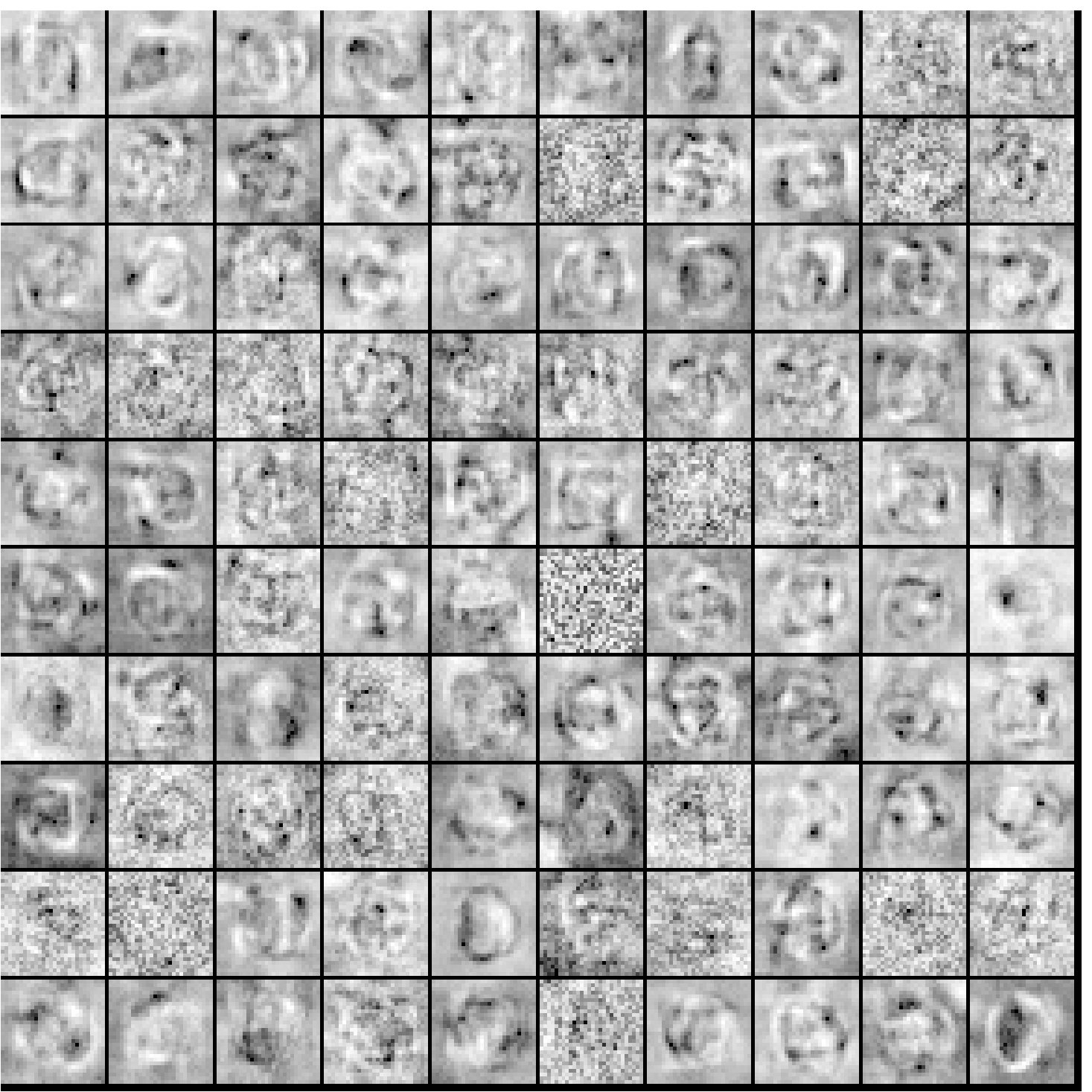}}
	 \end{center}
	\vspace{-1.0em}
        \caption{The 2D visualization of 100 randomly chosen weights after the DAE pretraining for each domain pairs from the Office image set. 
The white, gray, and black pixels in each box indicate the high-positive, close-to-zero or zero, and high-negative values of a particular connection weight.
The zero-masking noise is used with 30\% destruction.
}
	\label{fig:pretrained_weights}
\end{figure}

In this experiment, we ran both the \emph{unsupervised} and \emph{semi-supervised} adaptation setting for all domain pairs.
In addition, we also investigated the effect of DAE pretraining that precedes the NN and DaNN supervised training with respect to the performance.
The DAE pretraining will slightly change Step 1 of Algorithm~\ref{alg:dann}.
We denoted these models as \textbf{DAE + NN} and \textbf{DAE + DaNN}.
Examples of the pretrained weights are depicted in Figure~\ref{fig:pretrained_weights}.
The complete accuracy rates on the Office raw pixels for all domain pairs are presented in Table~\ref{tab:raw_results}.
\vspace{-1.0em}

\vspace{-3.0em}
\begin{table}[!htb]
\centering
	\caption{The performances on the Office dataset ($A$ : \emph{amazon}, $W$ : \emph{webcam}, $D$ : \emph{dslr}) using the raw pixels as inputs.}
	\scalebox{0.9}{
	\begin{tabular}{ c || c | c || c | c || c | c}
		\hline
		Methods & $A \rightarrow W$ & $W \rightarrow A$& $A \rightarrow D$ & $D \rightarrow A$  &  $W \rightarrow D$ & $D \rightarrow W$\\ \hline \hline
		\multicolumn{7}{c}{Unsupervised Setting} \\ \hline
		L-SVM & $14.9 \pm 0.0$ & $14.7 \pm 0.0$ & $19.1\pm 0.0$ & $13.7 \pm 0.0$ & $36.0 \pm 0.0$ & $40.3 \pm 0.0$ \\ \hline
		PCA + L-SVM & $20.3 \pm 0.0$ & $18.1 \pm 0.0$ & $16.9 \pm 0.0$ & $17.4 \pm 0.0$ & $40.4 \pm 0.0$ & $37.0 \pm 0.0$ \\ \hline
		GFK~\citep{Gong:2012aa} & $21.4 \pm 0.0$ & $15.0 \pm 0.0$ & $30.2 \pm 0.0$ & $13.8 \pm 0.0$ & $69.1 \pm 0.0$ & $65.0 \pm 0.0$ \\ \hline
		TSC~\citep{Long:2013aa} & $22.3 \pm 1.0$ & $15.7 \pm 1.1$ & $25.6 \pm 1.6$ & $19.6 \pm 0.7$ & \underline{$\mathbf{74.1\pm 1.9}$} & $67.5 \pm 1.5$ \\ \hline
		NN & $29.2 \pm 0.6$ & $17.0 \pm 0.3$ & $32.5 \pm 0.7$ & $15.0 \pm 0.3$ & $63.7 \pm 0.0$ & $57.3 \pm 0.0$ \\ \hline
		DAE + NN & $32.5 \pm 0.2$ &  $18.7 \pm 0.0$ & $37.8 \pm 0.2$ & $17.4 \pm 0.0$& $72.1 \pm 0.0$ & $65.9 \pm 0.0$  \\ \hline
		\textbf{DaNN} & $34.1 \pm 0.3$ & $21.2 \pm 0.2$  & $34.0 \pm 0.8$  & $20.1 \pm 0.5$ & $64.4 \pm 0.0$ & $62.0\pm 0.0$  \\ \hline
		\textbf{DAE + DaNN} & \underline{$\mathbf{35.0 \pm 0.2}$} & \underline{$\mathbf{23.1 \pm 0.0}$}  & \underline{$\mathbf{39.4 \pm 0.3}$}   & \underline{$\mathbf{22.5 \pm 0.0}$} & \underline{$\mathbf{74.3 \pm 0.0}$} & \underline{$\mathbf{70.5 \pm 0.0}$}  \\ \hline
		\multicolumn{7}{c}{Semi-supervised Setting} \\ \hline
		L-SVM & $18.9 \pm 0.0$   & $29.0 \pm 0.0$ & $25.2 \pm 0.0$  & $35.2 \pm 0.0$& $45.7 \pm 0.0$ & $52.5 \pm 0.0$  \\ \hline
		PCA + L-SVM & $20.8 \pm 0.0$ & $31.0 \pm 0.0$& $25.6 \pm 0.0$& $35.1 \pm 0.0$& $50.4 \pm 0.0$ & $50.2 \pm 0.0$\\ \hline
		GFK~\citep{Gong:2012aa} &$47.9 \pm 0.0$ &$33.1 \pm 0.0$ & $52.0 \pm 0.0$ & $31.8 \pm 0.0$ &$80.3 \pm 0.0$ & \underline{$\mathbf{74.7 \pm 0.0}$} \\ \hline
		TSC~\citep{Long:2013aa} &$42.4 \pm 2.1$ & $34.1 \pm 0.8$ & $49.3 \pm 2.2$ &$36.4 \pm 0.9$ & $76.3 \pm 1.4$ & $71.1 \pm 1.1$\\ \hline
		NN & $48.7 \pm 0.3$  & $34.5 \pm 0.3$ &  $52.8 \pm 0.6 $ & $36.2 \pm 0.4$ & $75.6 \pm 0.1$ & $67.2 \pm 0.0$ \\ \hline
		DAE + NN & $52.8 \pm 0.1$ & $36.8 \pm 0.0$ &$57.5 \pm 0.1$  & $36.5\pm 0.0$ & \underline{$\mathbf{83.5 \pm 0.0}$}& $69.4 \pm 0.0$ \\ \hline
		\textbf{DaNN} & $51.3 \pm 0.5$  & $36.6 \pm 0.4$ & $55.9 \pm 0.3$ & \underline{$\mathbf{37.9 \pm 0.3}$} & $78.0 \pm 0.2$ & $70.2 \pm 0.0$ \\ \hline
		\textbf{DAE + DaNN} &\underline{ $\mathbf{53.6 \pm 0.2}$}  &  \underline{$\mathbf{37.3 \pm 0.0}$}& \underline{$\mathbf{59.9 \pm 0.1}$} & \underline{$\mathbf{38.2 \pm 0.0}$} &  \underline{$\mathbf{83.5 \pm 0.0}$}& $71.2 \pm 0.0$ \\ \hline
	\end{tabular}
	}
	\label{tab:raw_results}
\end{table}
%In the case of using raw pixels as inputs, we also carry out the denoising auto-encoder (DAE) learning as the pre-training stage for both NN and DaNN. 
%We would like to investigate the effect of DAE pretraining on all unlabeled images of the Office data set with respect to the final performance of NN and DaNN.

It is clear that our DaNN always provides accuracy improvements in all domain pairs compared to the SVM-based baselines and the NN model.
In other words, the MMD regularization indeed improves the performance of neural networks.
Compared to TSC that also employs the MMD regularization in the unsupervised training stage, our DaNN performs better in most cases.
However, TSC can match the DaNN performance on \emph{webcam}-\emph{dslr} couples, which has lower level mismatch than the other couples.
This indicates that the utilization of the MMD regularization in the supervised training might gain more adaptation ability than that in the unsupervised training for pairs with more difficult mismatches to solve.

The DAE pretraining applied to NN and DaNN indeed improves the performances for all couples of domains.
The improvements are quite significant for several cases, especially for \emph{webcam}-\emph{dslr} couples.
In general, the DAE pretraining also produces more stable models in the sense of resulting in lower standard deviations over 30 independent runs.
Furthermore, the combination of DAE pretraining and DaNN performs best among other methods in these experiments in almost all cases.
In the sense of qualitative analysis, as can be seen in Figure~\ref{fig:pretrained_weights}, the DAE pretraining captures more distinctive ``filters`` from local blob detectors to object parts detectors, especially when the \emph{amazon} images are included.
This effect is somewhat consistent with what was found in the initial DAE work~\citep{Vincent:2010aa} suggesting that the DAE pretraining provides more useful neural network representations.

In the semi-supervised setting, the performance trend is somewhat similar to the unsupervised setting.
However, the performance discrepancies between NN and DaNN here becomes smaller than those in the unsupervised setting.
This outcome also holds for the case of the DAE pretraining.
This suggests that both the MMD regularization and DAE pretraining might be less impactful when some labeled images from the target domain can be acquired.

\vspace{-1.0em}
\subsection*{\emph{In-domain} Setting}
One may ask whether the domain adaptation results shown in Table~\ref{tab:raw_results} are reasonable compared to the standard learning setting.
We refer this standard setting to as the \emph{in-domain} setting, where the training and test samples come from the same domain.
The \emph{in-domain} performance can be considered as a reference that indicates the effectiveness of domain adaptation models in dealing with the domain mismatch.

We investigated the \emph{in-domain} performances of non-domain adaptive models described in Section~\ref{sec:ex_setup}, i.e., L-SVM, PCA+L-SVM, and NN on raw pixels of the Office images.
For each domain, we conducted $10$-fold cross-validation. 
%where the data set is divided into ten subsets of size $n/10$, nine subsets as the training set and one as the test set. 
%The evaluation is then repeated 10 times for each different configuration of training and test sets, and the results are averaged.
The complete \emph{in-domain} results in terms of the mean and standard deviation are shown in Table~\ref{tab:indomain}.
In general, we can see that the best \emph{in-domain} model is the NN model on both training and test images.

\begin{table}[!htb]
\centering
	\caption{The \emph{in-domain} performances on the Office data set using 10-fold cross-validation on each domain.}
	\scalebox{0.9}{
	\begin{tabular}{ c || c | c || c | c || c | c}
		\hline
		\multirow{2}{*}{Methods} &\multicolumn{2}{c ||}{\emph{amazon}} & \multicolumn{2}{c||}{\emph{webcam}}  &\multicolumn{2}{c}{\emph{dslr}} \\ \cline{2-7}
				& Training & Test & Training & Test & Training & Test \\ \hline \hline
		L-SVM & 99.0 $\pm$ 0.3 & 52.0 $\pm$ 4.6 & 100.0 $\pm$ 0.0 & 57.7 $\pm$ 13.9 & 100.0 $\pm$ 0.0 & 51.0 $\pm$ 14.1 \\
		PCA+L-SVM & 64.4 $\pm$ 0.8 & 60.6 $\pm$ 6.4 & 72.0 $\pm$ 1.5 & 62.8 $\pm$ 8.7 & 75.6 $\pm$ 2.1 & 55.2 $\pm$ 13.1 \\
		NN & 99.3 $\pm$ 0.1 & \underline{\textbf{74.2 $\pm$ 3.2}} & 100.0 $\pm$ 0.0 & \underline{\textbf{87.2 $\pm$ 5.4}} & 100.0 $\pm$ 0.0 &\underline{\textbf{ 77.9 $\pm$ 8.8}} \\ \hline
	\end{tabular}
	\label{tab:indomain}
	}
\end{table}

%PCA+L-SVM performs better than L-SVM on test sets, but worse on training sets.
%This shows that the use of PCA to reduce the dimensionality fails to cope with the \emph{bias-variance} trade-off.
%We also note that the standard deviations of the test accuracies are relatively high for all methods, which might indicate that the image variation within a particular domain in the Office set is high enough.
In comparison to the domain adaptation results, the highest \emph{in-domain} accuracies are better than the results with domain mismatches when the \emph{amazon} or \emph{webcam} are used as the target sets (see the highest accuracy rates in column $D \rightarrow A$ and $D \rightarrow W$ on Table~\ref{tab:raw_results}).
This indicates that a better domain adaptation model might be necessary to overcome those mismatches.
However, this is not the case for the \emph{dslr} as the target set where the \emph{in-domain} accuracy is even lower than the best domain adaptation result on $W \rightarrow D$ pair.
Knowing the facts that the \emph{webcam} and \emph{dslr} images are quite similar and the \emph{webcam} set has more images,
this shows that the domain adaptation indeed helps to produce a better object recognition model for this kind of setting.
%The most interesting outcome here is that the best \emph{in-domain} accuracies are, in general, still higher than the best performances with the domain mismatch when the \emph{amazon} or \emph{webcam} are used as the target sets (see the highest accuracy rates in column $D \rightarrow A$ and $D \rightarrow W$ on Table~\ref{tab:raw_results}).
%However, this is not case for $W \rightarrow D$ pair.
%This means that we might not need a better domain adaptation model any more for $W \rightarrow D$ pair since the upper bound has been surpassed.

\vspace{-1.0em}
\section{Conclusions and Future Work} 
This paper aimed to reduce the domain mismatch problem in object recognition using a simple neural network model, which we refer to as the Domain Adaptive Neural Network (DaNN).
In this work, we utilized the MMD measure as a regularization in the supervised back-propagation training.
This regularization encouraged the hidden layer representation distributions to be similar to each other.
We demonstrated that the DaNN performs well on the Office image set, especially on raw image pixels as inputs.
Furthermore, the DaNN preceded by the denoising auto-encoder (DAE) pretraining has better performance compared to SVM-based baselines, GFK~\citep{Gong:2012aa}, and TSC~\citep{Long:2013aa} on the Office image set~\citep{Saenko:2010aa} in almost all domain pairs.

Despite the effectiveness of the MMD regularization, there are still many aspects that can be further improved.
We have seen that the performance on raw pixels, which is a main concern in representation learning approach, is still not as good as that on SURF features.
We note that good models that perform well without any preceding handcrafted feature extractors are preferable to reduce complexity.
A better model on raw pixels might be achieved by using deeper neural network layers with a similar strategy since deep architectures have brought some successes in many applications in recent years~\citep{Bengio:2013ab}.
Our initial work using a standard deep neural network with the DAE pretraining, which is not shown here due to page limit, suggested that  deeper representations do not always improve the performance against the domain mismatch.

In addition, a study on the kernel choice for computing MMD regarding to the domain adaptation problem might be worth addressing.
We assumed that the universal Gaussian kernel function can detect any underlying distribution mismatches in the Office data set, which might be not true.
A better understanding about the relationship between a kernel function and a particular image mismatch, e.g., background, lighting, affine transformation changes, would induce a great impact in this field of research.
%However, the improvement is not significantly high for a few cases.
%It may be due to the Gaussian kernel parameters that was not optimally set for those cases.

\newpage
\vspace{-2cm}
\bibliographystyle{abbrvnat}
{\small\bibliography{../../Literatures/phdbib.bib}}

\begin{thebibliography}{21}
\providecommand{\natexlab}[1]{#1}
\providecommand{\url}[1]{\texttt{#1}}
\expandafter\ifx\csname urlstyle\endcsname\relax
  \providecommand{\doi}[1]{doi: #1}\else
  \providecommand{\doi}{doi: \begingroup \urlstyle{rm}\Url}\fi

\bibitem[Baktashmotlagh et~al.(2013)Baktashmotlagh, Harandi, Lovell, and
  Salzmann]{Baktashmotlagh:2013aa}
M.~Baktashmotlagh, M.~T. Harandi, B.~C. Lovell, and M.~Salzmann.
\newblock Unsupervised domain adaptation by domain invariant projection.
\newblock In \emph{Proceedings of International Conference on Computer Vision},
  pages 769--776, 2013.

\bibitem[Bay et~al.(2008)Bay, Tuytelaars, and Gool]{Bay:2008aa}
H.~Bay, T.~Tuytelaars, and L.~V. Gool.
\newblock Surf: Speeded up robust features.
\newblock \emph{Computer Vision and Image Understanding (CVIU)}, 110\penalty0
  (3):\penalty0 346--359, 2008.

\bibitem[Bengio(2013)]{Bengio:2013ab}
Y.~Bengio.
\newblock Deep learning of representations: Looking forward.
\newblock In \emph{Statistical Language and Speech Processing}, volume 7978 of
  \emph{Lecture Notes in Computer Science}, pages 1--37. Springer, 2013.

\bibitem[Bengio et~al.(2007)Bengio, Lamblin, Popovici, and
  Larochelle]{Bengio:2007aa}
Y.~Bengio, P.~Lamblin, D.~Popovici, and H.~Larochelle.
\newblock Greedy layer-wise training of deep networks.
\newblock In \emph{Advances in Neural Information Processing Systems (NIPS)},
  volume~19, page 153, 2007.

\bibitem[Bengio et~al.(2012)Bengio, Courville, and Vincent]{Bengio:2012ab}
Y.~Bengio, A.~C. Courville, and P.~Vincent.
\newblock Representation learning: A review and new perspectives.
\newblock \emph{Computing Research Repository}, abs/1206.5538, 2012.

\bibitem[Borgwardt et~al.(2006)Borgwardt, Gretton, Rasch, Kriegel,
  Sch\"{o}lkopf, and Smola]{Borgwardt:2006aa}
K.~M. Borgwardt, A.~Gretton, M.~J. Rasch, H.-P. Kriegel, B.~Sch\"{o}lkopf, and
  A.~J. Smola.
\newblock Integrating structured biological data by kernel maximum mean
  discrepancy.
\newblock \emph{Bioinformatics}, 22\penalty0 (14):\penalty0 e49--e57, 2006.

\bibitem[Cortes and Vapnik(1995)]{Cortes:1995aa}
C.~Cortes and V.~N. Vapnik.
\newblock {Support-Vector Networks}.
\newblock \emph{Machine Learning}, 20\penalty0 (3):\penalty0 273--297, 1995.

\bibitem[Daum\'{e}-III(2009)]{Daume-III:2009aa}
H.~Daum\'{e}-III.
\newblock Frustratingly easy domain adaptation.
\newblock \emph{CoRR}, abs/0907.1815, 2009.

\bibitem[Erhan et~al.(2010)Erhan, Bengio, Courville, Manzagol, and
  Vincent]{Erhan:2010aa}
D.~Erhan, Y.~Bengio, A.~Courville, P.-A. Manzagol, and P.~Vincent.
\newblock Why does unsupervised pre-training help deep learning?
\newblock \emph{Journal of Machine Learning Research}, 11:\penalty0 625--660,
  2010.

\bibitem[Glorot et~al.(2011)Glorot, Bordes, and Bengio]{Glorot:2011ab}
X.~Glorot, A.~Bordes, and Y.~Bengio.
\newblock Deep sparse rectifier neural network.
\newblock In \emph{Proceedings of the 14th International Conference on
  Artificial Intelligence and Statistics (AISTATS)}, pages 315--323, 2011.

\bibitem[Gong et~al.(2012)Gong, Shi, Sha, and Grauman]{Gong:2012aa}
B.~Gong, Y.~Shi, F.~Sha, and K.~Grauman.
\newblock Geodesic flow kernel for unsupervised domain adaptation.
\newblock In \emph{Proceedings of IEEE Conference on Computer Vision and
  Pattern Recognition (CVPR)}, pages 2066--2073, 2012.

\bibitem[Gopalan et~al.(2011)Gopalan, Li, and Chellapa]{Gopalan:2011aa}
R.~Gopalan, R.~Li, and R.~Chellapa.
\newblock Domain adaptation for object recognition: An unsupervised approach.
\newblock In \emph{IEEE International Conference on Computer Vision}, pages
  999--1006, 2011.

\bibitem[Gretton et~al.(2012)Gretton, Borgwardt, Rasch, Sch\''{o}lkopf, and
  Smola]{Gretton:2012aa}
A.~Gretton, K.~M. Borgwardt, M.~J. Rasch, B.~Sch\''{o}lkopf, and A.~Smola.
\newblock A kernel two-sample test.
\newblock \emph{Journal of Machine Learning Research}, pages 723--773, 2012.

\bibitem[Hinton et~al.(2012)Hinton, Srivastava, Krizhevsky, Sutskever, and
  Salakhutdinov]{Hinton:2012aa}
G.~E. Hinton, N.~Srivastava, A.~Krizhevsky, I.~Sutskever, and R.~Salakhutdinov.
\newblock Improving neural networks by preventing co-adaptation of feature
  detectors.
\newblock \emph{CoRR}, abs/1207.0580, 2012.

\bibitem[Long et~al.(2013)Long, Ding, Wang, Sun, Guo, and Yu]{Long:2013aa}
M.~Long, G.~Ding, J.~Wang, J.~Sun, Y.~Guo, and P.~S. Yu.
\newblock Transfer sparse coding for robust image representation.
\newblock In \emph{Proceedings of IEEE Conference on Computer Vision and
  Pattern Recognition (CVPR)}, pages 404--414, 2013.

\bibitem[Nair and Hinton(2010)]{Nair:2010aa}
V.~Nair and G.~E. Hinton.
\newblock Rectified linear units improve restricted boltzmann machines.
\newblock In \emph{Proceedings of the 27th International Conference on Machine
  Learning (ICML)}, 2010.

\bibitem[Pan and Yang(2010)]{Pan:2010aa}
S.~J. Pan and Q.~Yang.
\newblock A survey on transfer learning.
\newblock \emph{IEEE Transactions on Knowledge and Data Engineering},
  22\penalty0 (10):\penalty0 1345--1359, 2010.

\bibitem[Pan et~al.(2009)Pan, Tsang, Kwok, and Yang]{Pan:2009aa}
S.~J. Pan, I.~W. Tsang, J.~T. Kwok, and Q.~Yang.
\newblock Domain adaptation via transfer component analysis.
\newblock In \emph{Proceedings of the 21st International Joint Conference on
  Artificial Intelligence (IJCAI)}, pages 1187--1192, 2009.

\bibitem[Saenko et~al.(2010)Saenko, Kulis, Fritz, and Darrell]{Saenko:2010aa}
K.~Saenko, B.~Kulis, M.~Fritz, and T.~Darrell.
\newblock Adapting visual cateogry models to new domains.
\newblock In \emph{ECCV}, pages 213--226, 2010.

\bibitem[Steinwart(2002)]{Steinwart:2002aa}
I.~Steinwart.
\newblock On the influence of the kernel on the consistency of support vector
  machines.
\newblock \emph{Journal of Machine Learning Research}, 2:\penalty0 67--93,
  2002.

\bibitem[Vincent et~al.(2010)Vincent, Larochelle, Lajoie, Bengio, and
  Manzagol]{Vincent:2010aa}
P.~Vincent, H.~Larochelle, I.~Lajoie, Y.~Bengio, and P.-A. Manzagol.
\newblock Stacked denoising autoencoders: Learning useful representations in a
  deep network with a local denoising criterion.
\newblock \emph{Journal of Machine Learning Research}, 11:\penalty0 3371--3408,
  2010.

\end{thebibliography}

\end{document}